\documentclass{bmvc2k}

\usepackage{graphicx} 
\usepackage{hyperref}

\usepackage{titlesec}
\usepackage[normalem]{ulem}
\usepackage{amsmath}
\usepackage{amsfonts} 
\usepackage{enumitem}
\usepackage{multirow}
\usepackage[table,xcdraw]{xcolor}
\usepackage{booktabs}
\usepackage{fancyhdr}
\usepackage{sectsty}

\fancypagestyle{firstpage}{%
  \fancyhf{} 
  \fancyfoot[L]{$^\dagger$ Both authors contributed equally}
}

\definecolor{myforestgreen}{RGB}{34, 200, 34}

\newcommand{\stdvuno}[1]{\footnotesize{\color{black}(#1)} {\color{black}}}
\newcommand{\stdvun}[1]{\footnotesize\textcolor{black}{(#1)} \textcolor{myforestgreen}{$\uparrow$}}
\newcommand{\stdvunnn}[1]{\footnotesize\textcolor{black}{(#1)} \textcolor{myforestgreen}{$\uparrow\uparrow$}}
\newcommand{\stdvur}[1]{\footnotesize\textcolor{black}{(#1)} \textcolor{red}{$\downarrow$}}
\newcommand{\stdvunu}[1]{\footnotesize\textcolor{black}{(\underline{#1})} \textcolor{myforestgreen}{$\uparrow$}}
\newcommand{\stdvunnnu}[1]{\footnotesize\textcolor{black}{(\underline{#1})} \textcolor{myforestgreen}{$\uparrow\uparrow$}}
\newcommand{\stdvuru}[1]{\footnotesize\textcolor{black}{(\underline{#1})} \textcolor{red}{$\downarrow$}}
\newcommand{\stdvueq}[1]{\footnotesize\textcolor{black}{(\underline{#1})} \textcolor{black}{$\approx$}}

\title{FLARE up your data: Diffusion-based Augmentation Method in Astronomical Imaging}

\addauthor{Mohammed Talha Alam\textbf{*}}{mohammed.alam@mbzuai.ac.ae}{1}
\addauthor{Raza Imam\textbf{*}}{raza.imam@mbzuai.ac.ae}{1}
\addauthor{Mohsen Guizani}{mohsen.guizani@mbzuai.ac.ae}{1}
\addauthor{Fakhri Karray}{fakhri.karray@mbzuai.ac.ae}{2}

\addinstitution{
 Mohamed bin Zayed University of Artificial Intelligence,\\
 Abu Dhabi,\\
 United Arab Emirates
}
\addinstitution{
 University of Waterloo, \\
 Department of Electrical \& Computer Engineering,\\
 Canada
}

\runninghead{Alam \& Imam et al.}{FLARE up your data}

\begin{document}

\maketitle

\begin{abstract}
The intersection of Astronomy and AI encounters significant challenges related to issues such as noisy backgrounds, lower resolution (LR), and the intricate process of filtering and archiving images from advanced telescopes like the James Webb. 
Given the dispersion of raw images in feature space, we have proposed a \textit{two-stage augmentation framework} entitled as \textbf{FLARE} based on \underline{f}eature \underline{l}earning and \underline{a}ugmented \underline{r}esolution \underline{e}nhancement. We first apply lower (LR) to higher resolution (HR) conversion followed by standard augmentations. Secondly, we integrate a diffusion approach to synthetically generate samples using class-concatenated prompts. By merging these two stages using weighted percentiles, we realign the feature space distribution, enabling a classification model to establish a distinct decision boundary and achieve superior generalization on various in-domain and out-of-domain tasks.
We conducted experiments on several downstream cosmos datasets and on our optimally distributed \textbf{SpaceNet} dataset across 8-class fine-grained and 4-class macro classification tasks. FLARE attains the highest performance gain of 20.78\% for fine-grained tasks compared to similar baselines, while across different classification models, FLARE shows a consistent increment of an average of +15\%.
This outcome underscores the effectiveness of the FLARE method in enhancing the precision of image classification, ultimately bolstering the reliability of astronomical research outcomes. 
Our code and SpaceNet dataset is available at \href{https://github.com/Razaimam45/PlanetX_Dxb}{\textit{https://github.com/Razaimam45/PlanetX\_Dxb}}.

\end{abstract}

\section{Introduction}

In the age of technological advancement, telescopes such as James Webb \cite{kalirai2018scientific}, LSST \cite{hendel2019machine} and IFUs \cite{henneken2009sao} are producing vast amounts of data nightly, amounting to multiple terabytes, as they gather information on various cosmological phenomena. Managing this exponential increase in data complexity necessitates the development of automated tools within the field of astronomy to identify, analyze, and categorize celestial objects \cite{karypidou2021computer}.
With this requisite,
numerous complexities arise due to factors like varying object sizes, high contrast, low signal-to-noise ratios, noisy backgrounds, and diverse orbital scenarios \cite{aldahoul2022localization}. Current methods \cite{garcia2021lspnet}, primarily employing convolutional neural networks, aims to classify images featuring space objects against intricate backgrounds. Nevertheless, these approaches occasionally struggle to effectively highlight the objects within the images, resulting in classification errors and reduced accuracy \cite{alzubaidi2021review}. Additionally, there are several challenges in collecting and filtering images from the James Webb Telescope \cite{kalirai2018scientific}. These challenges include enhancing initial image quality during data collection, improving image categorization and organization for adequate filtering, and facilitating efficient data archiving. 
Moreover, existing datasets, whether from academic sources or public repositories like Kaggle or NASA, often consist of LR images without structured categorization. For instance, the Galaxy Zoo dataset \cite{masters2010galaxy} focuses on a limited range of galaxy classes. These limitations limit the potential for comprehensive and high-quality research in this field, highlighting the need for well-structured datasets and efficient data archiving processes.

\begin{figure}[t]
    \centering
    \includegraphics[width=0.80\textwidth]{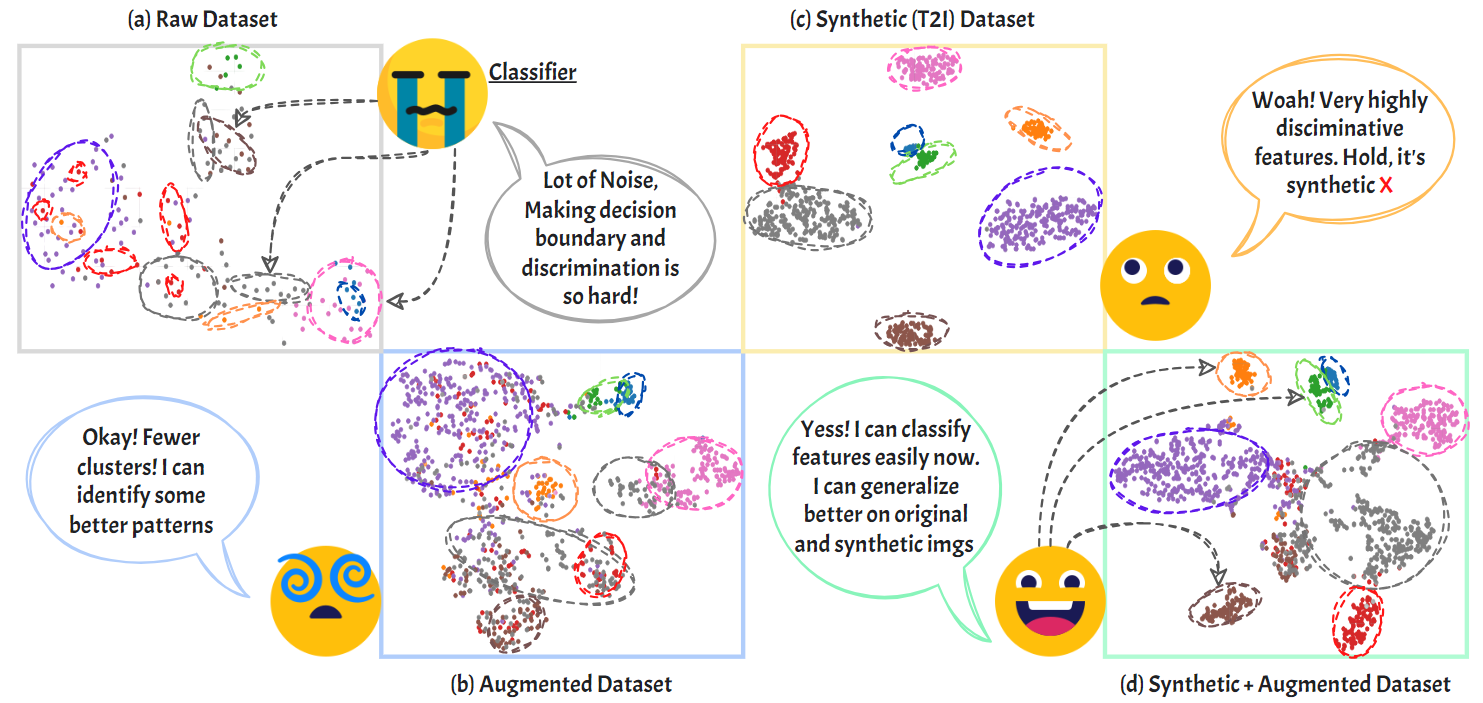}
    \caption{The combination of traditional augmentation and synthetic samples created through diffusion benefits in terms of harmonizing feature representations, achieving higher classification performance.}
    \label{fig:overview}
\end{figure}

\noindent The raw cosmos dataset, similar to other existing datasets \cite{kaggle-dataset}, presents challenges due to its wide range of classes with varying distributions, resulting in a broader distribution and, subsequently, lower model accuracy.
Moreover, by preprocessing our raw dataset, we face several challenges of overlapping features in the feature space. To overcome these challenges, we initially employ HR followed by standard data augmentation. We employed SwinIR \cite{liang2021swinir} to enhance LR images, elevating them to HR quality and ensuring the dataset consistently contains high-quality space images. While comparatively improving distribution shifts, augmentation still falls short of optimal test generalization. As a solution, we introduce a stable diffusion approach, generating samples based on curated prompts. These diffusion-based samples undergo image restoration and integration with the raw HR dataset, resulting in a narrower data distribution for each class. This process enhances \textit{discriminative features} and promotes better class separation, as shown in Figure \ref{fig:overview}.

\noindent With the explosion of generative models, there has been a boom in training on synthetic data, which can lead to \textit{model collapse} \cite{shumailov2023curse}, \textit{i.e.}, deviating from optimal performance on real-world problems. \cite{seddik2024bad} have shown that model collapse cannot be avoided when models are trained solely on synthetic data. To address this issue, \cite{seddik2024bad} has demonstrated that mixing both real and synthetic data can help mitigate practical issues like model collapse.
Specifically, we utilize UniDiffuser \cite{bao2023one} to create synthetic samples by employing curated textual prompts concatenated with class labels. We then combine the synthetic dataset with the augmented HR dataset to generate an optimally distributed dataset, selecting augmentations from each dataset based on weighted percentiles. We achieved a more robust dataset by merging generated samples with the original HR dataset. This led to more focused class distributions, improving the classifier's ability to distinguish features and enhance classification performance.
In summary, our technical contributions are as follows,
\begin{itemize}
    \item We introduce \textbf{FLARE}, a diffusion-based framework that initially converts LR data into HR, and generates an optimally combined data by mixing real and synthetic distribution, consistently enhancing test performance by +20\% compared to simply augmented data.
    \item The LR-to-HR module of FLARE employs SwinIR \cite{liang2021swinir} to enhance LR images, ensuring high-quality data both aesthetically and in terms of improved performance.
    \item Our optimally distributed dataset via FLARE, entitled \textbf{SpaceNet}, comprising approximately 12,900 samples, is the first in the astronomy domain to consist of hierarchically structured HR images that achieve up to 85\% test accuracy on an 8-class fine-grained classification task.
\end{itemize}

\section{Related Work}

\textbf{Astronomy and Computer Vision.} Recent advancements in the fusion of astronomy and computer vision have yielded notable progress. Hendel et al. \cite{aldahoul2022localization} introduced SCUDS, a novel machine-vision technique utilizing the SCMS algorithm to automate the classification of tidal debris structures, shedding light on galaxy assembly histories. In meteor detection, Al-Owais et al. \cite{al2023meteor} utilized YOLOv3 and YOLOv4 object detection algorithms to distinguish meteors from non-meteor objects, achieving impressive recall and accuracy scores while enhancing monitoring efficiency. Additionally, Shirasuna et al. \cite{shirasuna2023optimized} proposed an optimized training approach with an attention mechanism for robust meteor detection, particularly beneficial with limited data.

\noindent \textbf{Upscaling and Restoration.} 
Image restoration models have made a significant impact on astronomy and image enhancement domains. Zhang et al. \cite{zhang2020residual} introduced the Residual Dense Network (RDN), effectively addressing the challenge of utilizing hierarchical features from low-quality images by integrating local and global features, achieving outstanding results in various restoration tasks. Zamir et al. \cite{zamir2022restormer} proposed Restormer, an efficient Transformer model tailored for handling computational complexity in high-resolution image tasks like deraining, motion deblurring, defocus deblurring, and denoising. In unsupervised image restoration, Poirier-Ginter \cite{poirier2023robust} presented a robust StyleGAN-based approach capable of handling different degradation levels and types of image degradation, employing a 3-phase latent space extension and a conservative optimizer to achieve realistic results compared to diffusion-based methods.

\noindent \textbf{Classification Models.} Pretrained and fine-tuned deep learning models, such as ResNet, EfficientNet, Inception, GoogleNet, ViT, and DenseNet, exhibit versatility in classifying diverse objects across astronomy and other domains \cite{szegedy2015going,he2016deep,tan2019efficientnet,dosovitskiy2020image,imam2023enhancing,imam2023optimizing}. Becker et al. \cite{becker2021cnn} addressed challenges in radio galaxy morphology classification, investigating potential overfitting in CNNs trained with limited datasets and evaluating various architectures. Fluke et al. \cite{fluke2020surveying} provided a comprehensive survey of AI applications in astronomy, covering classification, regression, clustering, forecasting, generation, discovery, and insights. Fielding et al. \cite{fielding2021comparison} conducted a comparative analysis of deep learning architectures for optical galaxy morphology classification, highlighting DenseNet-121's superior accuracy and training efficiency.


\section{Methodology}

\subsection{Preliminaries}
\label{sec:appendix_preliminaries}
\textbf{Image Restoration.}
In our image restoration module of FLARE, we employ the Swin Transformer (SwinIR) for enhancing low-quality input images $I_{LR} \in \mathbb{R}^{H \times W \times C_{i n}}$ where \textit{H} represents the image height, \textit{W} signifies the image width, and \textit{$C_{in}$} corresponds to the number of input channels \cite{liang2021swinir}. This begins by extracting shallow features $F_{0} \in \mathbb{R}^{H \times W \times C}$ through a 3x3 convolutional layer, which performs visual processing and feature mapping. Subsequently, deep features $F_{DF} \in \mathbb{R}^{H \times W \times C}$ are derived from $F_{0}$ using the deep feature extraction module. During image reconstruction, the high-quality image $I_{HR}$ is reconstructed by combining shallow and deep features via the reconstruction module $H_{REC}$ as
$I_{HR} = H_{REC}(F_0 + F_{DF})$.
The image restoration process is defined as,
\begin{align}
\label{eq:swinir}
I_{HR} = SwinIR(I_{LR}) = \mathcal{F}_{swin}(\mathcal{E}_{swin}(I_{LR})) + Loss 
\end{align}
where $\mathcal{E}_{swin}(.)$ refers to the encoder that extracts features from the input image, while $\mathcal{F}_{swin}(.)$ represents the decoder responsible for reconstructing the output image from these features. The $Loss$ function quantifies the difference between the input and output images produced by the SwinIR model as \(Loss(I_{LR}, I_{HR})\).

\noindent \textbf{Diffusion Models.}
Diffusion models offer multi-modal image generation through two distinct stages: the forward noise process and the reverse denoising process. In the forward process, a sequence $x_{1}, x_{2}, ..., x_{T}$ is generated from a starting point $x_{0}$, drawn from the distribution $p(x_{0})$, by iteratively adding noise. This process relies on the equation
$q\left(\mathbf{x}_t \mid \mathbf{x}_0\right) = N\left(\mathbf{x}_t ; \alpha_t \mathbf{x}_0, \sigma_t^2 \mathbf{I}\right) \nonumber$.
Here, each step introduces a noise component $\epsilon$, sampled from the standard normal distribution. Conversely, the denoising process models the transition from $x_{t}$ to $x_{t-1}$ through the conditional probability $p_\theta\left(\mathbf{x}_{t-1} \mid \mathbf{x}_t\right)$. In this stage, the predicted statistics, $\hat{\mu}_\theta\left(\mathbf{x}_t\right), \hat{\Sigma}_\theta\left(\mathbf{x}_t\right)$, are determined, guided by a learnable parameter $\theta$. Optimization occurs through a loss function $\ell_{\text{simple}}^t(\theta)$ \cite{rombach2022high}
quantifying the difference between actual and predicted noise, leveraging a learnable neural network.
The trained neural network, known as the predictor, can then generate an image $\hat{\mathbf{x}}_0$. 
We employed the UniDiffuser diffusion model \cite{bao2023one}for handling various data types, allowing effective generation of image-text and image-image pairs without added complexity. UniDiffuser, designed for text-to-image tasks, is represented as,
\begin{align}
    \label{eq:unidiffuser}
    \text{UniDiffuser}(prompt) = \textit{F}(\mathcal{T}(prompt), \mathcal{G}(image))
\end{align}
 where $\mathcal{T}$\textit{(prompt)} and $\mathcal{G}$\textit{(image)} stand for the text encoder and image generator, respectively. The fusion module \textit{F} efficiently combines features from the text encoder and image generator, focusing on producing perceptually realistic results.

\begin{figure}[t]
    \centering
    \includegraphics[width=1\textwidth]{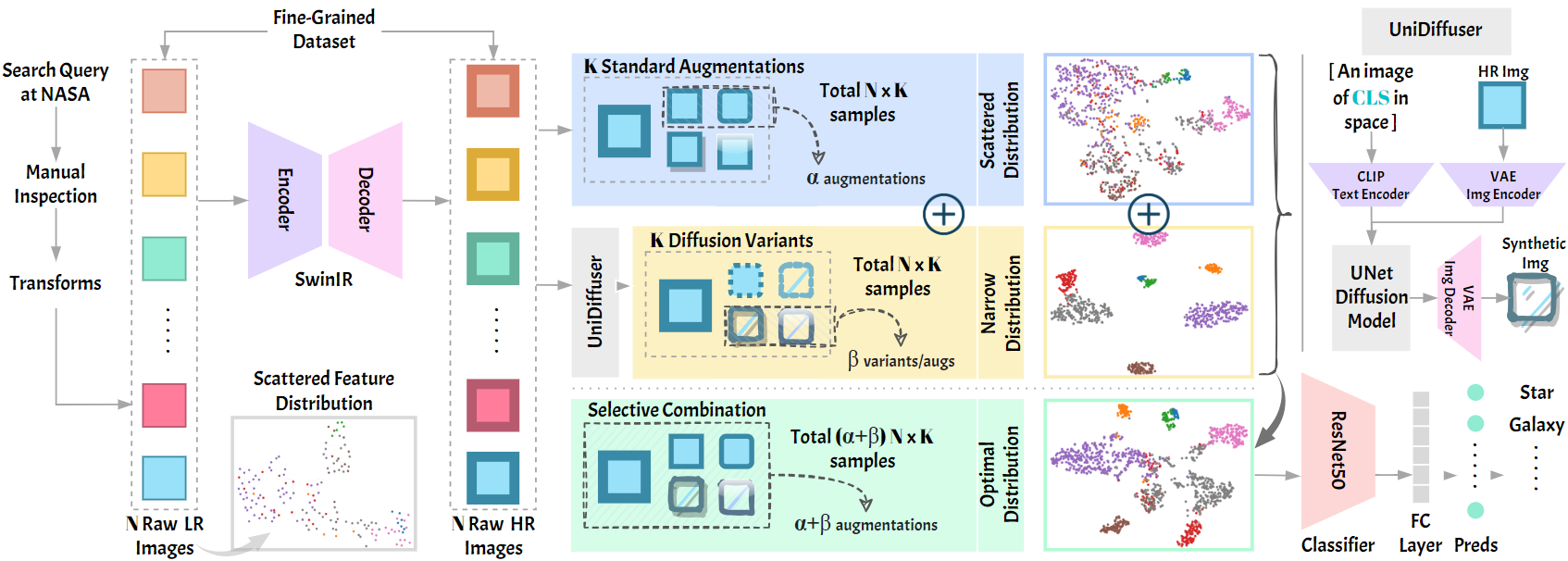}
    \caption{Our proposed methodology, FLARE: We upscale a raw dataset to high resolution using SwinIR. Next, we apply standard augmentation techniques. Then, we create synthetic samples by combining class-concatenated prompts with UniDiffuser. We then select relevant augmentations and combine them based on weighted percentiles. The resulting optimally distributed dataset is then fed into a classifier for enhanced classification.}
    \label{fig:method}
\end{figure}

\subsection{Optimal Cosmic Classification}

\textbf{High-Resolution Dataset.} 
Raw images of outer space typically contain a lot of noise and variability, making it difficult for a classifier to learn meaningful features. Features for different classes might overlap because there's insufficient discrimination in the raw pixel values.
Without proper feature extraction, the model may struggle to distinguish between classes \cite{belle2021principles}. 
Hence, we first convert the LR images to HR. Given an extracted raw dataset with LR images, $D^{LR}$ consists of $n$ samples, where each sample is represented as $(x_i^{LR}, y_i)$. These samples include a set of $x_i$ images and corresponding $y_i$ labels. We perform a LR to HR conversion using SwinIR. $D^{LR}$ is transformed into $D^{HR}$  such that $\forall (x_i^{LR}, y_i)$ $\epsilon$ $D^{LR}$ is input to SwinIR to attain $D^{HR}$ with samples as $(x_i^{HR}, y_i)$, $\forall$ $i$ ranging from $1$ to $n$. Specifically using Eq. \ref{eq:swinir}, 
\begin{align}
    \label{eq:lr2hr}
    D^{HR} &= \mathcal{F}_{swin}(\mathcal{E}_{swin}(D^{LR}))
    = \mathcal{F}_{swin}(\mathcal{E}_{swin}(\{(x_i^{LR}, y_i)\}) \nonumber \\
    &= \{(x_i^{HR}, y_i)\} \hspace{0.5cm} \text{$\forall$ $i$} = \{1,~ \ldots~ n\}
\end{align}
where $\mathcal{E}_{swin}(.)$ and $\mathcal{F}_{swin}(.)$ are the image encoder and image decoder of the SwinIR, respectively.

\noindent \textbf{Augmentation and Diffusion.}
Data augmentations like rotation, scaling, and cropping can help the model generalize better by providing more varied examples \cite{perez2017effectiveness_raw_aug_lr}.
Augmentation introduces additional variability and helps the model learn to handle different variations within the same class \cite{mumuni2022data}.
Figure \ref{fig:overview} showing better clustering suggests that augmentation helps the model identify common patterns within each class compared to just raw samples.
Thus, building upon the high-resolution dataset $D^{HR}$, we apply $k-1$ different augmentations (RandomFlip and ColorJitter) to each sample in $D^{HR}$ (from Eq. \ref{eq:lr2hr}) to create an augmented dataset $D^{HR}_{Aug}$, denoted as,
\begin{align}
    D^{HR}_{Aug} 
    = D^{HR} + \mathcal{A}_{1}(D^{HR}) + \mathcal{A}_{2}(D^{HR}) + \ldots +\mathcal{A}_{k-1}(D^{HR}) = D^{HR} + \sum_{i=1}^{k-1} \mathcal{A}_{i}(D^{HR})
\end{align}
where $\mathcal{A}_{i}$ is the $i^{th}$ type of augmentation on the input dataset. For example, $\mathcal{A}_{1}$ may indicate color jitter while $\mathcal{A}_{2}$ could represent random flip of the input samples. Following $k-1$ different augmentations, $D^{HR}_{Aug}$ has a total of $n\text{*}k$ samples, where $n$ is the number of samples and $k$ is the number of augmentations.

\noindent Synthetic images generated using diffusion \cite{weng2023diffusion}, \cite{corvi2023detection} or similar methods \cite{shamsolmoali2021image} can produce highly discriminative features, which can be designed to be more distinct and less noisy than raw data, making class separation easier \cite{tan2022robust}. Thus, we generate a new diffusion-based dataset $D^{HR}_{T2I}$ followed by its augmentation. For this, we initially use the class ($CLS$) based prompts like "A realistic image of \underline{$CLS$} in space" as prompt (text) initialization to input Unidiffuser (Eq. \ref{eq:unidiffuser}) with Image-to-Text generation task such that: $D_{T2I}$ $\leftarrow$ \text{UniDiffuser}$\langle$Prompts concatenated with $CLS$ $\rangle$.
\begin{align}
    \label{eq:T2I}
    D_{T2I} &= \mathcal{F}_{diff}(\mathcal{E}_{diff}(P_{y_1}, P_{y_2}, P_{y_3}, \ldots, P_{y_n}) = \mathcal{F}_{diff}(\mathcal{E}_{diff}(P_{y_i})) \nonumber \\
    &= \{(x_1^{T2I}, y_1), \ldots, (x_n^{T2I}, y_n)\}
    = \{(x_i^{T2I}, y_i)\} \hspace{0.5cm} \text{$\forall$ $i$} = \{1,~ \ldots~ n\}
\end{align}
where $\mathcal{E}_{diff}(.)$ and $\mathcal{F}_{diff}(.)$ are text encoder and decoder of the UniDiffuser respectively, while $P_{y_i}$ is the prompt produced using the label of $i^{th}$ class, which is then input to the encoder. This text-to-image (T2I) diffusion-based dataset $D_{T2I}$ consists of $n$ samples. Now, we generate further $k-1$ variations of the samples in $D_{T2I}$ dataset. Generating variations of T2I samples is a form of data augmentation, which is generated as follows,
\begin{align}
    \label{eq:I2I}
    D_{I2I} 
    = D_{T2I} + \mathcal{V}_{1}(D_{T2I}) + \mathcal{V}_{2}(D_{T2I}) + \ldots +\mathcal{V}_{k-1}(D_{T2I}) = D_{T2I} + \sum_{i=1}^{k-1} \mathcal{V}_{i}(D_{T2I})
\end{align}
where $\mathcal{V}_{i}(.)$ is the I2I (Image-to-Image) module of UniDiffsuer which generates a variation of the input image with seed $i$. This augmented version of $D_{T2I}$, \textit{i.e.}, $D_{I2I}$, now consists of $n\text{*}k$ samples, equalling that of $D^{HR}_{Aug}$. Using Eq. \ref{eq:lr2hr}, we generate HR version of $D_{I2I}$ as $D^{HR}_{T2I}$, both having high resolution and equal number of samples.

\noindent \textbf{Preserving optimal distribution.}
Combining augmented raw images with synthetic variations offers both variety and discriminative features. The classifier benefits from increased diversity and clear separation achieved by synthetic data. We have two augmented versions of the HR dataset, $D^{HR}_{Aug}$ and $D^{HR}_{T2I}$ , which are combined in a weighted manner. This weighted combination selects a percentile ratio of samples based on the augmentation types. In simpler terms, it selects $\alpha$ and $\beta$ augmentations out of the $k$ augmentations from the $D^{HR}_{Aug}$ and $D^{HR}_{T2I}$ datasets, as follows,
\begin{align}
    \label{eq:selection}
    \tilde{D}^{HR}_{Aug} &= \mathcal{S}(\alpha, k, D^{HR}_{Aug}) \nonumber \\
    \text{and}\quad \tilde{D}^{HR}_{T2I} &= \mathcal{S}(\beta, k, D^{HR}_{T2I}) \\
    \text{s.t.}\quad 0 < \alpha &\leq 1 \nonumber; 0 < \beta \leq 1 \nonumber
\end{align}
where $\mathcal{S}(.)$ is the selection function that selects the $\alpha$ percentile of augmentations out of the $k$ total augmentations for ${D}^{HR}_{Aug}$. Following the selection of optimal augmentations, we attain $\tilde{D}^{HR}_{Aug}$ and $\tilde{D}^{HR}_{T2I}$ which consists of $n\text{*}(\alpha\cdot k)$ and $n\text{*}(\beta\cdot k)$ number of samples respectively. We concatenate these two augmented datasets to obtain a optimally distributed dataset as,
\begin{align}
    \label{eq:add_alp_beta}
    \tilde{D}^{HR} &= \tilde{D}^{HR}_{Aug} + \tilde{D}^{HR}_{T2I}
\end{align}
where $\tilde{D}^{HR}$ consists of $(\alpha+\beta)\cdot n\text{*}k$   number of samples. \\

\noindent \textbf{Enhanced Classification.} The resulting combined dataset $\tilde{D}^{HR}$, incorporating both augmented and synthetically generated data, increases the variance within the training samples, thus endowing the \textit{FLARE model} (i.e., model trained on $\tilde{D}^{HR}$) with a more diverse range of inputs to learn from. This augmentation substantially bolsters the generalizability of FLARE model as it adapts to a wider in-domain and out-of-domain distribution, encapsulating both real and synthetic data. This intricate interplay ensures that the model not only harnesses the discriminative attributes offered by synthetic samples but also attains the ability to navigate and comprehend the inherent variations within a single class. In effect, the combined dataset contributes to increased class separation, ultimately facilitating the establishment of more refined decision boundaries.


\begin{figure}[!b]
\vspace{-0.25cm}

    \centering
    \includegraphics[width=\textwidth]{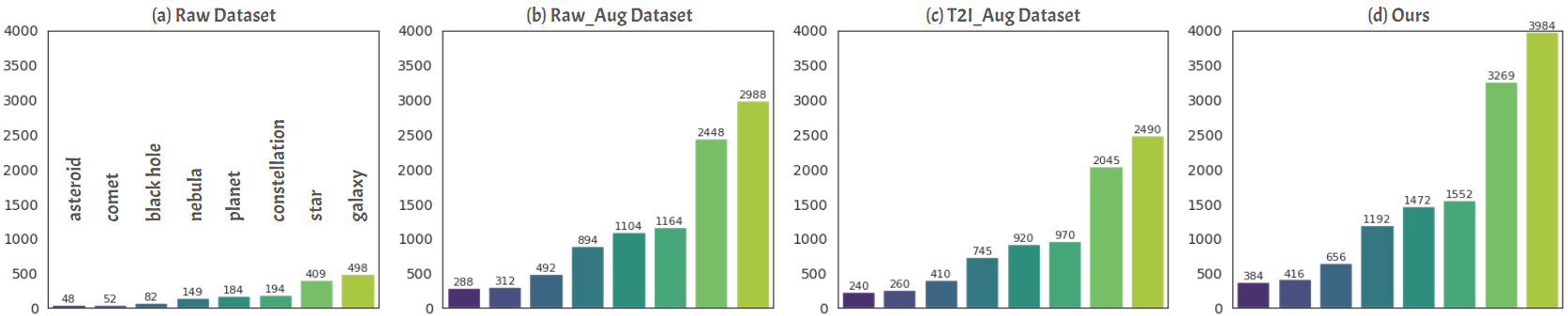}
    \caption{The original raw dataset (Raw\_Aug), when transformed into our combined dataset using the FLARE approach, results in 7.8$\times$ increase in the number of samples. \textbf{Ours} represent the proposed \textbf{SpaceNet} dataset.}
    \label{fig:cls_dist}
\end{figure}

\section{Experiments and Results}
\subsection{Dataset}
Data collection for FLARE involved web scraping from NASA's official website using refined classname-based search queries. Employing Python libraries BeautifulSoup and Chrome Driver, we obtained around 2,500 images classified into eight distinct astronomical categories: planets, galaxies, asteroids, nebulae, comets, black holes, stars, and constellations. Manual curation was necessary to eliminate non-conforming images like logos and artistic renditions, resulting in the initial raw dataset $D^{LR}$ comprising 1,616 images across eight fine-grained classes. Four umbrella classes were then created based on these fine-grained classes: "Astronomical Patterns," "Celestial Bodies," "Cosmic Phenomena," and "Stellar Objects," facilitating a hierarchical analytical approach. Additionally, we generated a synthetic dataset $D_{T2I}$ using UniDiffuser with $class$-concatenated text prompts, resulting in 8,080 synthetic images, matching the number of raw augmented images in $D^{HR}_{Aug}$. Using FLARE, we obtain our SpaceNet dataset which represents \textit{in-domain} distribution as shown in Figure \ref{fig:cls_dist}. For \textit{out-of-domain} datasets, representing \textit{downstream tasks}, we utilize several existing datasets including GalaxyZoo \cite{masters2010galaxy}, Space \cite{abhikalp_srivastava_space_images_category}, and Spiral \cite{altruistic_emphasis_spiral_galaxies}. 
Further experimental details, including implementation, baselines, classifiers, and metrics, are provided in Appendix \ref{sec:appendix_exp}.

{\renewcommand{\arraystretch}{1.2}
\begin{table}[b!]

\centering
\caption{Quantitative assessment of upscaled images in terms of PSNR and MS-SSIM. Average PSNR values ranging above 25 represents high image restoration quality \cite{liang2021swinir}.}
\resizebox{\textwidth}{!}{%
\begin{tabular}{l|lllllllll}
\hline
\textbf{Classes\_avg} & \textbf{asteroid} & \textbf{comet} & \textbf{black hole} & \textbf{nebula} & \textbf{planet} & \textbf{constellation} & \textbf{star} & \textbf{galaxy} & \textbf{AVG} \\ \hline
\rowcolor[HTML]{f6fbff} 
\textbf{PSNR} & 30.57 & 29.23 & 25.06 & 25.12 & 31.02 & 24.95 & 25.48 & 25.16 & \textbf{27.07} \\
\rowcolor[HTML]{f6fbff} 
\textbf{MS-SSIM} & 0.78 & 0.76 & 0.67 & 0.68 & 0.81 & 0.67 & 0.65 & 0.67 & \textbf{0.71} \\ \hline
\end{tabular}
}
\label{tab:psnr_mmssim}
\end{table}

\begin{figure}[t]
    \centering
    \includegraphics[width=0.90\textwidth]{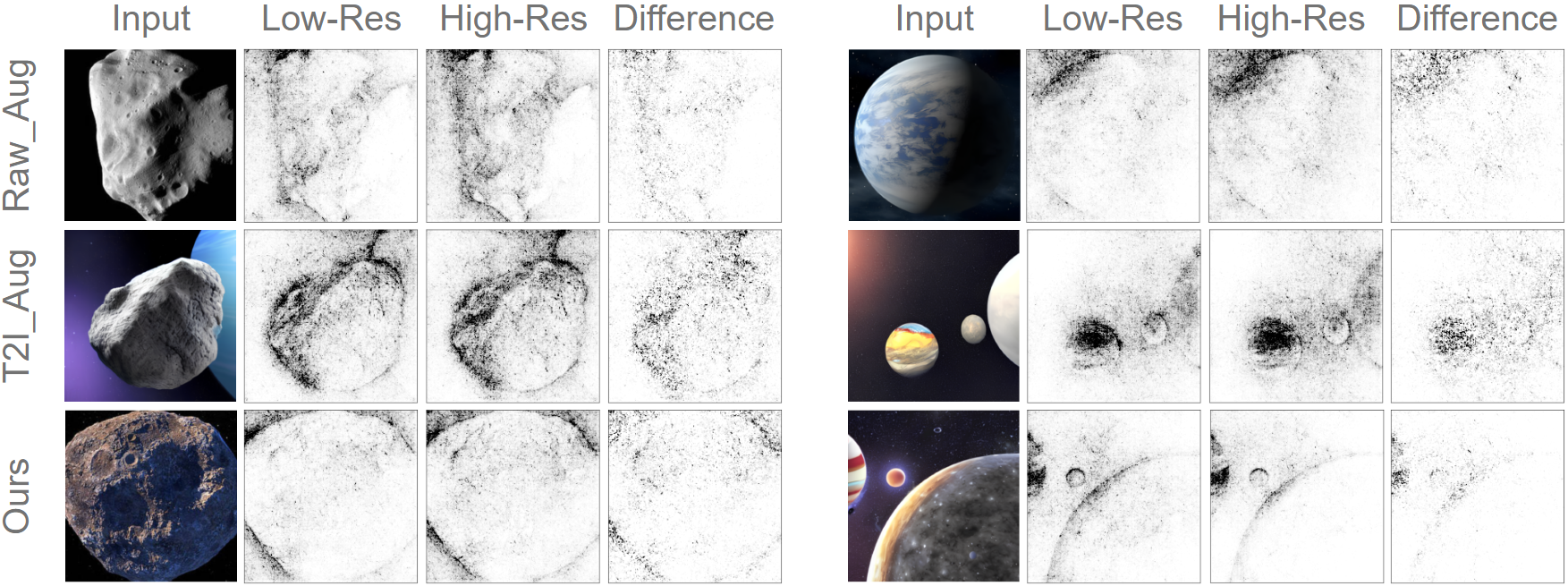}
    \caption{Integrated Gradient for LR inputs, HR inputs, and their difference across different methods, illustrating visual relationship between the model's predictions and the extracted features. LR-to-HR module of FLARE helps to embed discriminative features in input space.}
    \label{fig:integrated_grads}
\vspace{-0.10cm}
\end{figure}

\begin{figure}[t]
    \centering
    \includegraphics[width=0.95\textwidth]{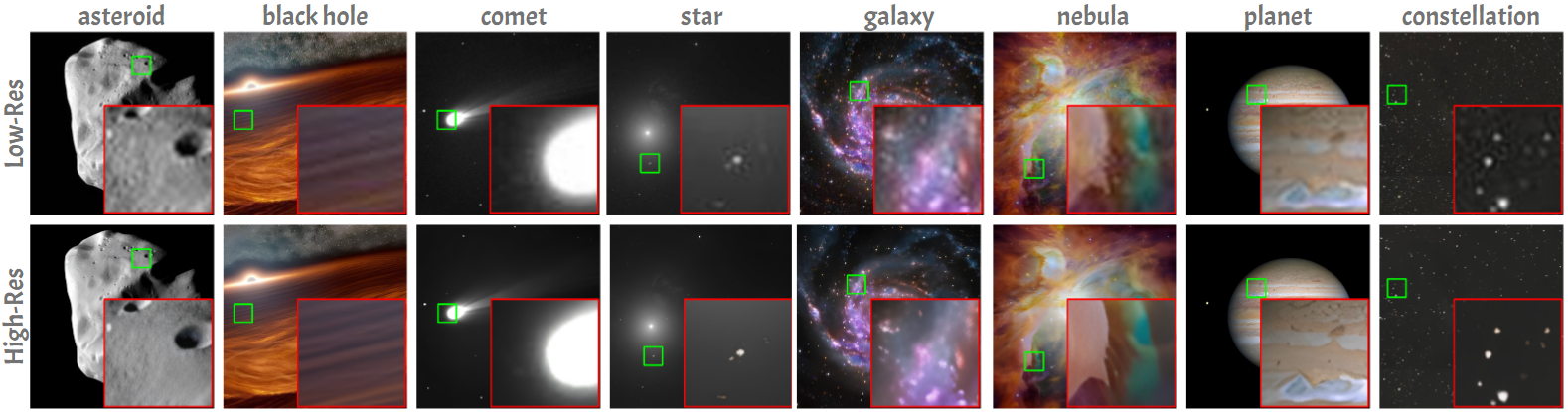}
    \caption{Visual comparison of upscaled noise restoration examples across 8 classes}
    \label{fig:patch_zoom}
\end{figure}

 \subsection{Low Resolution \textit{to} High Resolution}
The initial evaluation of models trained on LR images showed average accuracies of around 60.61\% for fine-grained and 69.81\% for macro classes (Table \ref{tab:acc_f1}). Augmentations to LR images improved performance by approximately 5\% to 6\%. However, due to limitations like high noise and overlapping class features, we applied SwinIR to generate $D^{HR}_{Aug}$. Evaluation revealed an average PSNR of 27.07 and an average MS-SSIM of 0.71 across all fine-grained classes (Table \ref{tab:psnr_mmssim}). SwinIR effectively reduced noise and improved image clarity, as demonstrated visually and through integrated-gradients analysis (Figure \ref{fig:integrated_grads}). HR images exhibited sharper edges and more realistic textures, indicating their superiority for fine-grained classification tasks as shown in Figure \ref{fig:patch_zoom}. As a result, this method notably enhanced accuracy, with increases of approximately 9\% to 10\% across various classifiers for both macro and fine-grained classes (Table \ref{tab:acc_f1}). Assessments are conducted on a fixed test set of raw dataset $D^{LR}_{Aug}$.

{\renewcommand{\arraystretch}{1.2}
\begin{table}[!t]
\centering
\caption{Quantitative assessment of 4 classifiers across different methodologies for Macro and Fine-grained classes stating \textbf{in-domain} accuracy. \textbf{FLARE} indicate models trained on our \textbf{SpaceNet} dataset, where SpaceNet is combined with $\alpha=0.5$ and $\beta=1.0$ (Using Eq. \ref{eq:selection} and \ref{eq:add_alp_beta}). \textbf{Average} indicates average performance across all classifiers.}
\resizebox{\textwidth}{!}{%
\begin{tabular}{l|l|llll|l}
\hline
\rowcolor[HTML]{FFFFFF} 
\cellcolor[HTML]{FFFFFF} & \cellcolor[HTML]{FFFFFF} & \textbf{ResNet-50 \cite{he2016deep}} & \textbf{GoogleNet \cite{szegedy2015going}} & \textbf{DenseNet-121 \cite{huang2017densely}} & \textbf{ViT-B/16 \cite{dosovitskiy2020image}} & \textbf{Average}\\ \cmidrule(r){3-3} \cmidrule{4-4} \cmidrule(l){5-5} \cmidrule(l){6-6} \cmidrule(l){7-7}
\rowcolor[HTML]{FFFFFF} 
\multirow{-2}{*}{\cellcolor[HTML]{FFFFFF}\textbf{Data Type}} & \multirow{-2}{*}{\cellcolor[HTML]{FFFFFF}\textbf{Method}} & Accuracy & Accuracy & Accuracy & Accuracy & Accuracy \\ \hline
\rowcolor[HTML]{F2F2F2} 
\cellcolor[HTML]{FFFFFF} & Raw\_LR \cite{carrasco2019deep_raw_lr} \stdvuno{bs.} & 69.45\stdvuno{bs.} & 68.52\stdvuno{bs.} & 69.04\stdvuno{bs.} & 72.24\stdvuno{bs.} & 69.81\stdvuno{bs.} \\
\rowcolor[HTML]{f6fbff} 
\cellcolor[HTML]{FFFFFF} & Raw\_Aug\_LR \cite{perez2017effectiveness_raw_aug_lr} & 74.61\stdvun{5.16} & 75.03\stdvun{6.51} & 76.57\stdvun{7.53} & 76.57\stdvun{4.33} & 75.69\stdvun{5.88} \\
\rowcolor[HTML]{eff6fc} 
\cellcolor[HTML]{FFFFFF} & Raw\_Aug\_HR & 78.02\stdvun{8.57} & 77.50\stdvun{8.98} & 79.57\stdvun{10.53} & 80.91\stdvun{8.67} & 79.00\stdvun{9.19} \\
\rowcolor[HTML]{e6f1fa} 
\cellcolor[HTML]{FFFFFF} & T2I\_Aug\_HR & 80.12\stdvun{10.67} & 80.44\stdvun{11.92} & 80.44\stdvun{11.40} & 81.45\stdvun{9.21} & 80.61\stdvun{10.80}\\
\rowcolor[HTML]{DDEBF7} 
\multirow{-5}{*}{\cellcolor[HTML]{FFFFFF}Macro} & \textbf{FLARE} (Ours) & \textbf{87.69\stdvunnn{18.24}} & \textbf{84.80\stdvunnn{16.28}} & \textbf{85.74\stdvunnn{16.70}} & \textbf{86.44\stdvunnn{14.20}} & \textbf{86.16\stdvunnn{16.35}}\\ \hline
\rowcolor[HTML]{F2F2F2} 
\cellcolor[HTML]{FFFFFF} & Raw\_LR \cite{carrasco2019deep_raw_lr} \stdvuno{bs.} & 60.68\stdvuno{bs.} & 59.55\stdvuno{bs.} & 60.27\stdvuno{bs.} & 61.92\stdvuno{bs.} & 60.61\stdvuno{bs.} \\
\rowcolor[HTML]{f6fbff} 
\cellcolor[HTML]{FFFFFF} & Raw\_Aug\_LR \cite{perez2017effectiveness_raw_aug_lr} & 66.15\stdvun{5.47} & 67.18\stdvun{7.63} & 67.39\stdvun{7.12} & 68.21\stdvun{6.29} & 67.23\stdvun{6.62} \\
\rowcolor[HTML]{eff6fc} 
\cellcolor[HTML]{FFFFFF} & Raw\_Aug\_HR & 70.38\stdvun{9.70} & 70.07\stdvun{10.52} & 72.96\stdvun{12.69} & 72.03\stdvun{10.11} & 71.36\stdvun{10.75}\\
\rowcolor[HTML]{e6f1fa} 
\cellcolor[HTML]{FFFFFF} & T2I\_Aug\_HR & 77.16\stdvun{16.48} & 75.84\stdvun{16.29} & 76.85\stdvun{16.58} & 77.01\stdvun{15.09} & 76.72\stdvun{16.11}\\
\rowcolor[HTML]{DDEBF7} 
\multirow{-5}{*}{\cellcolor[HTML]{FFFFFF}Fine-grained} & \textbf{FLARE} (Ours) & \textbf{83.94\stdvunnn{23.26}} & \textbf{81.29\stdvunnn{21.74}} & \textbf{83.79\stdvunnn{23.52}} & \textbf{82.70\stdvunnn{20.78}} & \textbf{82.93}\stdvunnn{22.32}\\ \hline
\end{tabular}
}
\label{tab:acc_f1}
\vspace{-0.25cm}
\end{table}

\subsection{Raw \textit{vs} Diffusion Classification}
We synthesized the dataset \(D^{HR}_{T2I}\) using Eq. \ref{eq:T2I} and Eq. \ref{eq:I2I} and evaluated it for fine-grained classification. Models trained on \(D^{HR}_{T2I}\) average test accuracies of 80.61\% for macro and 76.72\% for fine-grained classification, marking substantial improvements of approximately +10\% and +16\%, respectively than the models trained on $D^{LR}_{Aug}$ dataset (Table \ref{tab:acc_f1}). These gains were consistent across various classification models.
This can be attributed to the diffusion model's effective alignment with the generated image distributions. Synthetic images from \(D^{HR}_{T2I}\) are designed to minimize noise and standardize features, resulting in more distinct, class-specific feature representations \cite{croitoru2023diffusion}. This reduction in intra-class variability and enhanced feature clarity is visible in the well-separated feature clusters shown in Figure \ref{fig:tsne_fg}. Thus, models trained on diffusion-based synthetic dataset \(D^{HR}_{T2I}\) consistently show improved in-domain classification performance but may fail to generalize against out-of-domain tasks.

\begin{figure}[!b]
\vspace{-0.5cm}

    \centering
    \includegraphics[width=\textwidth]{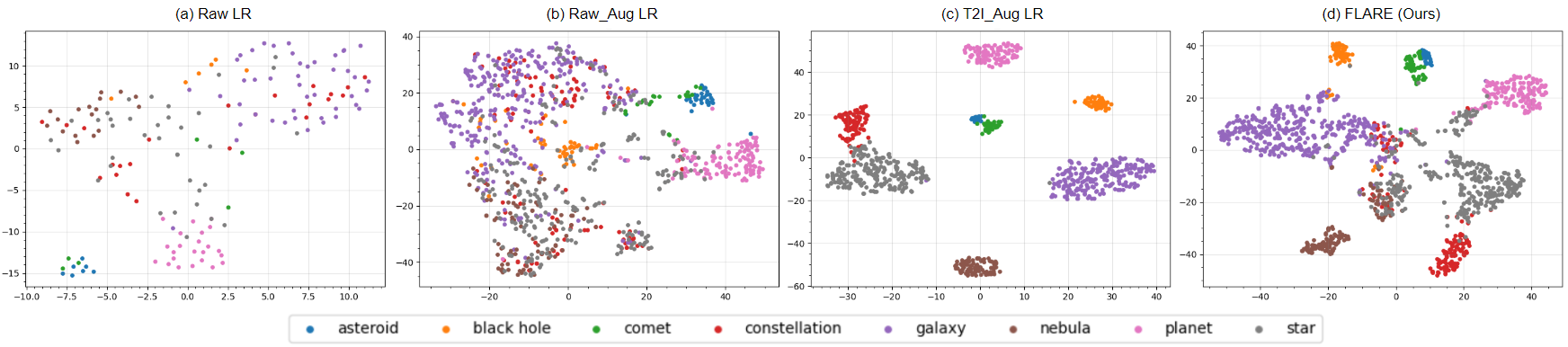}
    \caption{t-sne plot\textbf{s} representing the distribution of features across different dataset variants representing 8 Fine-grained classes.}
    \label{fig:tsne_fg}
\end{figure}

{\renewcommand{\arraystretch}{1.5}
\begin{table}[t]
\label{tab:downstream}
\caption{Quantitative results in terms of \textbf{out-of-domain} accuracy across different downstream tasks. This illustrates the generalization of pretrained backbones (\textit{i.e.}, models trained on \textbf{SpaceNet} (our) dataset) when tested on the relevant downstream datasets.}
\fontsize{10}{10}\selectfont
\resizebox{\textwidth}{!}{%
\begin{tabular}{l|l|llll|l} 

\hline
\rowcolor[HTML]{FFFFFF} 
\cellcolor[HTML]{FFFFFF} & \cellcolor[HTML]{FFFFFF} & \multicolumn{5}{c}{\cellcolor[HTML]{FFFFFF}\textbf{Downstream Dataset}} \\ \cmidrule{3-7}

\rowcolor[HTML]{FFFFFF} 
\multirow{-2}{*}{\cellcolor[HTML]{FFFFFF}\textbf{\begin{tabular}[c]{@{}l@{}}Pretrained \\ Backbone\end{tabular}}} & \multirow{-2}{*}{\cellcolor[HTML]{FFFFFF}\textbf{Method}} & \textbf{GalaxyZoo} \cite{masters2010galaxy} & \textbf{Space} \cite{abhikalp_srivastava_space_images_category} & \textbf{Spiral} \cite{altruistic_emphasis_spiral_galaxies} & \textbf{SpaceNet} & \textbf{Average} \\ \cmidrule(r){1-1} \cmidrule(lr){2-2} \cmidrule(r){3-3} \cmidrule{4-4} \cmidrule(l){5-5} \cmidrule(l){6-6} \cmidrule(l){7-7}

\rowcolor[HTML]{F2F2F2} 
\cellcolor[HTML]{FFFFFF} & Raw\_LR \cite{carrasco2019deep_raw_lr} \stdvun{bs.} & 33.91\stdvun{bs.} & 41.67\stdvun{bs.} & 83.10\stdvun{bs.} & 60.68\stdvun{bs.} & 54.84\stdvun{bs.} \\
\rowcolor[HTML]{F6FBFF} 
\cellcolor[HTML]{FFFFFF} & Raw\_Aug\_LR \cite{perez2017effectiveness_raw_aug_lr} & \textbf{72.75} \stdvun{38.84} & 41.67 \stdvueq{0.00} & 88.73 \stdvun{5.63} & 66.15 \stdvun{5.47} & 67.33 \stdvun{12.49} \\
\rowcolor[HTML]{DDEBF7} 
\multirow{-3}{*}{\cellcolor[HTML]{FFFFFF}ResNet-50 \cite{he2016deep}} & \textbf{FLARE} (Ours) & 71.76 \stdvur{37.25} & \textbf{59.38} \stdvun{17.71} & \textbf{95.77} \stdvun{12.67} & \textbf{83.94} \stdvun{23.26} & \textbf{77.71} \stdvunnn{22.87}\\ \hline

\rowcolor[HTML]{F2F2F2} 
\cellcolor[HTML]{FFFFFF} & Raw\_LR \cite{carrasco2019deep_raw_lr} \stdvun{bs.} & 53.88\stdvun{bs.} & 40.62\stdvun{bs.} & 84.51\stdvun{bs.} & 59.55\stdvun{bs.} & 59.64\stdvun{bs.} \\
\rowcolor[HTML]{F6FBFF} 
\cellcolor[HTML]{FFFFFF} & Raw\_Aug\_LR \cite{perez2017effectiveness_raw_aug_lr} & \textbf{66.46} \stdvun{12.58} & 45.83 \stdvun{5.21} & 81.69 \stdvur{2.82} & 67.18 \stdvun{7.63} & 65.29 \stdvun{5.65} \\
\rowcolor[HTML]{DDEBF7} 
\multirow{-3}{*}{\cellcolor[HTML]{FFFFFF}GoogleNet \cite{szegedy2015going}} & \textbf{FLARE} (Ours) & 65.84 \stdvur{11.96} & \textbf{56.25} \stdvun{15.63} & \textbf{92.96} \stdvun{8.45} & \textbf{81.29} \stdvun{21.74} & \textbf{74.09} \stdvunnn{14.45} \\ \hline

\rowcolor[HTML]{F2F2F2} 
\cellcolor[HTML]{FFFFFF} & Raw\_LR \cite{carrasco2019deep_raw_lr} \stdvun{bs.} & 37.73\stdvun{bs.} & 48.96\stdvun{bs.} & 80.28\stdvun{bs.} & 60.27\stdvun{bs.} & 56.81\stdvun{bs.} \\
\rowcolor[HTML]{F6FBFF} 
\cellcolor[HTML]{FFFFFF} & Raw\_Aug\_LR \cite{perez2017effectiveness_raw_aug_lr} & \textbf{70.16}  \stdvun{32.43} & 48.96 \stdvueq{0.00} & 87.32 \stdvun{7.04} & 67.39 \stdvun{7.12} & 68.46 \stdvun{11.65} \\
\rowcolor[HTML]{DDEBF7} 
\multirow{-3}{*}{\cellcolor[HTML]{FFFFFF}DenseNet-121 \cite{huang2017densely}} & \textbf{FLARE} (Ours) & 65.84 \stdvur{28.11} & \textbf{59.38} \stdvun{10.42} & \textbf{94.37} \stdvun{14.09} & \textbf{83.79} \stdvun{23.52} & \textbf{75.85} \stdvunnn{22.04} \\ \hline

\rowcolor[HTML]{F2F2F2} 
\cellcolor[HTML]{FFFFFF} & Raw\_LR \cite{carrasco2019deep_raw_lr} \stdvun{bs.} & 67.20\stdvun{bs.} & 45.83\stdvun{bs.} & 90.14\stdvun{bs.} & 61.92\stdvun{bs.} & 66.27\stdvun{bs.} \\
\rowcolor[HTML]{F6FBFF} 
\cellcolor[HTML]{FFFFFF} & Raw\_Aug\_LR \cite{perez2017effectiveness_raw_aug_lr} & \textbf{93.09} \stdvun{25.89} & 44.79 \stdvur{1.04} & 87.32 \stdvur{2.82} & 68.21 \stdvun{6.29} & 73.35 \stdvun{7.08}\\
\rowcolor[HTML]{DDEBF7} 
\multirow{-3}{*}{\cellcolor[HTML]{FFFFFF}ViT-B/16 \cite{dosovitskiy2020image}} & \textbf{FLARE} (Ours) & 83.72 \stdvur{16.52} & \textbf{55.21} \stdvun{9.38} & \textbf{98.59} \stdvun{8.45} & \textbf{82.70} \stdvun{20.78} & \textbf{80.06} \stdvunnn{13.79} \\ \hline

\end{tabular}}
\vspace{-0.25cm}
\end{table}}

\subsection{Distribution Shift}
The proposed framework FLARE incorporates the fusion of conventional augmented data and synthetically generated data in an optimal manner, in addition to conversion to higher quality, which is then used to train classification models. In the following sections, we discuss how distribution shifts occur across several phases of FLARE.\\

\noindent \textbf{Raw to Augmentation.} 
While data augmentation may introduce a degree of variability in model predictions, these consequences typically bode well for training more effective and reliable models. This is evident from Figure \ref{fig:tsne_fg}, which reveals that the features of the models trained on $D^{LR}$ \cite{perez2017effectiveness_raw_aug_lr} (Figure \ref{fig:tsne_fg}(a)) are dispersed across the feature space, lacking a distinct decision boundary and ultimately resulting in a lower fine-grained accuracy of 60.61\% and 54.84\% across in-domain and out-of-domain evaluations respectively. In contrast, models trained on $D^{LR}_{Aug}$ attains an average accuracy of 67.23\% and 67.33\% across in-domain and out-of-domain tasks respectively. This increase in accuracy following augmentation highlights the improved clustering in feature space, reflecting the benefits of augmentation ((Figure \ref{fig:tsne_fg}(b))). Nonetheless, achieving a clearer decision boundary remains a challenge.\\

\noindent \textbf{Synthetic Data.} 
Models trained on synthetic dataset $D^{HR}_{T2I}$ exhibits entirely distinct features for each class, as shown in Figure \ref{fig:tsne_fg} (c) and Figure \ref{fig:std_dev}. However, when evaluated on $D^{LR}$ (which represents raw real-world noisy data), such models encounter challenges in achieving robust generalization. This is primarily due to variations in the raw dataset that are not replicated in the synthetic samples $D^{HR}_{T2I}$. In other words, synthetic samples $D^{HR}_{T2I}$ may not precisely replicate the characteristics of real-world raw data $D^{LR}$, leading to nuanced differences in distribution, noise patterns, or other traits that impact the model's performance when dealing with genuine data. Nonetheless, the diversity and variance introduced by synthetic samples are harnessed to benefit our combined dataset-SpaceNet.\\

\begin{figure}[h]
    \centering
    \includegraphics[width=0.75\textwidth]{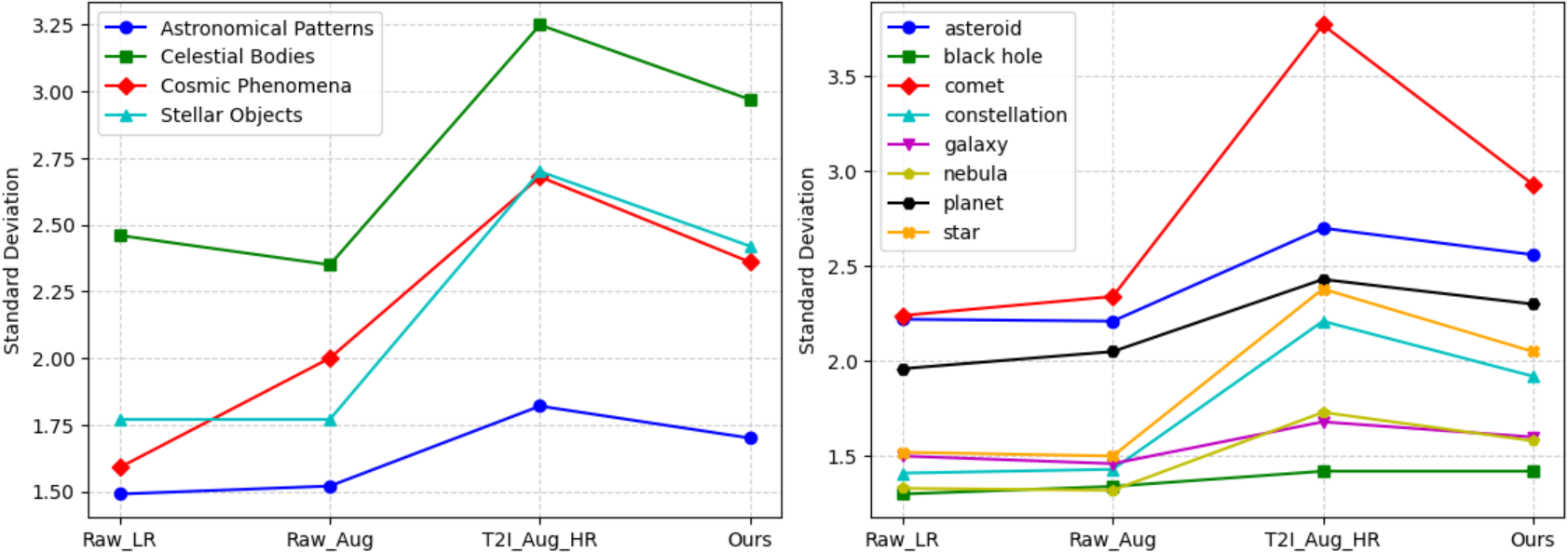}
    \caption{Standard Deviation across different datasets and classes. Ours (representing combined dataset) achieves optimal variance comparatively.}
    \label{fig:std_dev}
\end{figure}

\noindent \textbf{Combination of Augmented and Synthetic Data.} 
Our SpaceNet dataset $\tilde{D}^{HR}$, achieved through a weighted combination of samples from the augmented dataset $D^{HR}_{Aug}$ and the synthetic dataset $D^{HR}_{T2I}$ using a judiciously weighted percentile approach, relies on the strengths of both methodologies. As shown in Table \ref{tab:acc_f1}, FLARE models (\textit{i.e.}, classifiers trained on SpaceNet $\tilde{D}^{HR}$) exhibits an average improvement of 22.32\% and 22.87\% across in-domain and out-of-domain generalization, when compared to the models trained on raw dataset $D^{LR}$ during evaluation (See Table \ref{tab:acc_f1} and Table \ref{tab:downstream}). 
Similarly, in contrast to models trained on $D^{HR}_{Aug}$ and $D^{HR}_{T2I}$, FLARE achieves average accuracy increments of about +12\% and +6\%, respectively. 
Our FLARE framework reaches peak efficiency when weight percentiles are \textit{optimally} selected, favoring reliable augmentations. In our most successful scenario, we selected two augmentations, Color Jitter and Random Flip, alongside all samples from the diffusion-based synthetic dataset, as illustrated in Table \ref{tab:alpha_beta}. 
By blending the high variance of the diffusion-based synthetic dataset with the variance of the augmented dataset $\tilde{D}^{HR}$ using Eq. \ref{eq:add_alp_beta}, the combined dataset-SpaceNet $\tilde{D}^{HR}$  achieves an optimal distribution characterized by optimal variance and conducive feature distribution in feature space (See Figure \ref{fig:std_dev}). This diversity preserves distinct features across classes, leading to a well-defined decision boundary and enhancing classification accuracy.

{\renewcommand{\arraystretch}{1.7}
\begin{table}[t]
\fontsize{8}{8}\selectfont
\centering
\caption{Different values of $\alpha$ and $\beta$ for weighted combination on the Fine-grained SpaceNet using the ResNet-50 \cite{he2016deep} classifier. $\textbf{*}$=Best, Underline=w.r.t previous row comparison.}
\resizebox{0.8\textwidth}{!}{%
\begin{tabular}{llllllll}
\hline
\textbf{Method} & \textbf{$\alpha$} & \textbf{$\beta$} & \textbf{Accuracy} & \textbf{F1-Score} & \textbf{Precision} & \textbf{Recall} \\ \hline
\rowcolor[HTML]{F2F2F2} 
Raw\_LR \cite{carrasco2019deep_raw_lr} \stdvuno{bs.} & 0.00 & 0.00 & 60.68\stdvuno{bs.} & 55.80\stdvuno{bs.} & 64.56\stdvuno{bs.} & 60.68\stdvuno{bs.} \\
\rowcolor[HTML]{F2F2F2} 
Raw\_Aug\_HR & 1.00 & 0.00 & 70.38\stdvunu{9.7} & 70.53\stdvunu{14.73} & 71.03\stdvunu{6.47} & 70.38\stdvunu{9.7} \\ \hline
\rowcolor[HTML]{f6fbff} 
FLARE (Ours) & 0.50 & 0.50 & 75.34\stdvunu{4.96} & 75.52\stdvunu{4.99} & 75.85\stdvunu{4.82} & 75.34\stdvunu{4.96} \\
\rowcolor[HTML]{eff6fc} 
FLARE (Ours) & 0.50 & 0.75 & 79.27\stdvunu{3.93} & 79.08\stdvunu{3.56} & 79.27\stdvunu{3.42} & 79.04\stdvunu{3.7} \\
\rowcolor[HTML]{e6f1fa} 
FLARE (Ours) & 1.00 & 1.00 & 78.94\stdvuru{0.33} & 79.03\stdvuru{0.05} & 79.27\stdvueq{0.00} & 78.94\stdvuru{0.10} \\
\rowcolor[HTML]{DDEBF7} 
\textbf{FLARE*} (Ours) & \textbf{0.50} & \textbf{1.00} & \textbf{83.94}\stdvunnnu{5.00} & \textbf{83.58}\stdvunnnu{4.55} & \textbf{83.94}\stdvunnnu{4.67} & \textbf{83.94}\stdvunnnu{5.00} \\
\hline
\end{tabular}
}
\label{tab:alpha_beta}
\vspace{-0.25cm}
\end{table}}

\section{Conclusion}
This paper presents FLARE, a two-stage augmentation method for improving image classification accuracy in astronomy. By combining traditional and diffusion-based augmentation, FLARE effectively addresses challenges related to noisy backgrounds, lower resolution, and data filtering issues. Our approach consistently outperforms standard augmentation methods, particularly when utilizing high-resolution training samples. Our findings emphasize the effectiveness of FLARE in enhancing image classification precision, which can benefit space research reliability.
\noindent In the \textit{future}, we plan to address data imbalance issues to further enhance classification results. FLARE offers a cost-effective solution for astronomical research, streamlining data filtering and archiving processes and facilitating finer image detail extraction during initial data analysis stages.\\










\bibliography{references} 

\newpage
\appendix

\noindent\section{\begin{huge} \textbf{Appendix} \vspace{4mm} \end{huge}}







\subsection{Additional Experimental Details}
\label{sec:appendix_exp}


\noindent \textbf{Downstream Datasets.} In-domain tasks represent generalization performance when training and testing are done on the same dataset, while out-of-domain tasks represent performance when testing is done on a downstream dataset different from the one used for training.\\

\noindent \textbf{Baselines.} To assess the effectiveness of our proposed approach, we compare it against two groups of methods in astronomical classification. The first group involves classification performed directly on extracted images $D^{LR}$ (referred to as $Raw\_LR$), while the second group involves augmented versions of the raw images $D^{LR}_{Aug}$ (referred to as $Raw\_Aug\_LR$). 
For the $Raw\_LR$ group, our initial baseline is established by \cite{carrasco2019deep_raw_lr}. Moving to the $Raw\_Aug\_LR$ group, we consider \cite{perez2017effectiveness_raw_aug_lr} as our second baseline, as it demonstrates the effectiveness of data augmentation techniques in image classification.\\

\noindent \textbf{Classifiers.}
We utilized state-of-the-art pre-trained CNN models to train our custom cosmos dataset, including ResNet-50 \cite{he2016deep}, GoogleNet \cite{szegedy2015going}, DenseNet-121 \cite{huang2017densely}, and ViT-B/16 \cite{dosovitskiy2020image}, which were initially trained on the ImageNet dataset \cite{deng2009imagenet}. Our primary task involved classifying two distinct datasets: one for fine-grained data with 8 classes and the other for macro-level categories with 4 classes. Through these extensive experiments on various ConvNets, we aimed to assess how well these models could classify data in different situations, highlighting their flexibility and adaptability.\\

\noindent \textbf{Metrics.}
We use two essential metrics for assessing the quality of HR images derived from LR versions. The first metric, Peak Signal-to-Noise Ratio (PSNR) \cite{tanchenko2014visual}, measures image noise, with higher values indicating improved image quality. The second metric, Multi-Scale Structural Similarity (MS-SSIM) \cite{hore2010image}, evaluates structural and textural fidelity, with higher MS-SSIM values signifying better preservation of intricate details in HR images compared to LR. Furthermore, for classification models, we employed standard metrics, including Accuracy, F1-Score, Precision, and Recall, to assess the performance of the classifiers across all dataset variants \cite{alam2022s}.

\subsection{Additional Results}
\label{sec:appendix_results}

\begin{figure}[h]
    \centering
    \includegraphics[width=\textwidth]{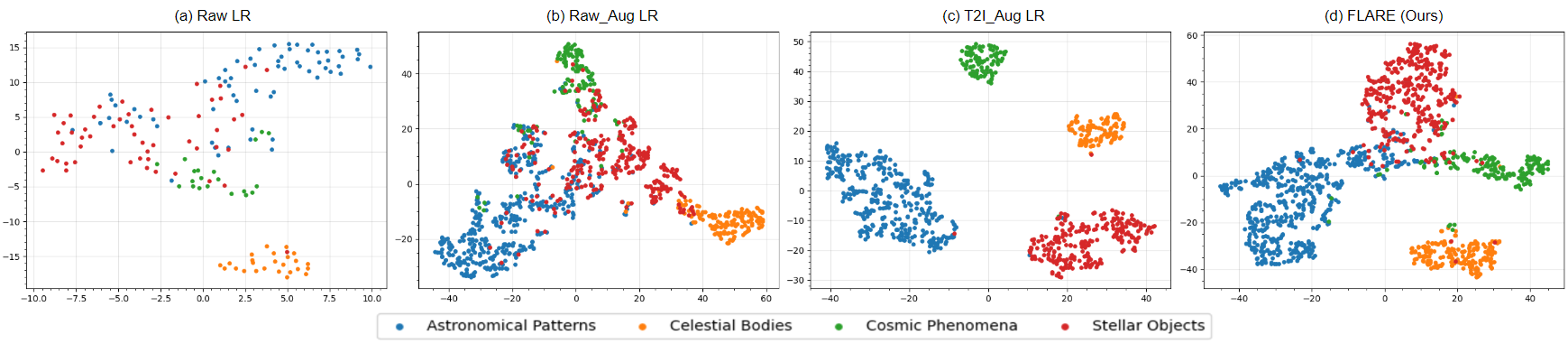}
    \caption{t-sne plot\textbf{s} representing the distribution of features across different dataset variants. representing 4 macro classes.}
    \label{fig:tsne_macro}
\end{figure}

{\renewcommand{\arraystretch}{1.3}
\begin{table}[h]
\centering
\caption{Quantitative assessment of 4 classifiers across different methodologies for Macro and Fine-grained classes stating \textbf{in-domain} F1-Scores. \textbf{FLARE} indicate models trained on our \textbf{SpaceNet} dataset, where SpaceNet is combined with $\alpha=0.5$ and $\beta=1.0$ (Using Eq. \ref{eq:selection} and \ref{eq:add_alp_beta}). \textbf{Average} indicates average performance across all classifiers.}
\resizebox{\textwidth}{!}{%
\begin{tabular}{l|l|llll|l}
\hline
\rowcolor[HTML]{FFFFFF} 
\cellcolor[HTML]{FFFFFF} & \cellcolor[HTML]{FFFFFF} & \textbf{ResNet-50 \cite{he2016deep}} & \textbf{GoogleNet \cite{szegedy2015going}} & \textbf{DenseNet-121 \cite{huang2017densely}} & \textbf{ViT-B/16 \cite{dosovitskiy2020image}} & \textbf{Average} \\ \cmidrule(r){3-3} \cmidrule{4-4} \cmidrule(l){5-5} \cmidrule(l){6-6} \cmidrule(l){7-7}

\rowcolor[HTML]{FFFFFF} 
\multirow{-2}{*}{\cellcolor[HTML]{FFFFFF}\textbf{Data Type}} & \multirow{-2}{*}{\cellcolor[HTML]{FFFFFF}\textbf{Method}} & F1-Score & F1-Score & F1-Score & F1-Score & F1-Score \\ \hline
\rowcolor[HTML]{F2F2F2} 
\cellcolor[HTML]{FFFFFF} & Raw\_LR \cite{carrasco2019deep_raw_lr} \stdvuno{bs.} & 68.83\stdvuno{bs.} & 67.38\stdvuno{bs.} & 67.92\stdvuno{bs.} & 72.00\stdvuno{bs.} & 69.03\stdvuno{bs.} \\
\rowcolor[HTML]{f6fbff} 
\cellcolor[HTML]{FFFFFF} & Raw\_Aug\_LR \cite{perez2017effectiveness_raw_aug_lr} & 74.57\stdvun{5.74} & 75.19\stdvun{7.81} & 76.04\stdvun{8.12} & 76.03\stdvun{4.03} & 75.45\stdvun{6.42} \\
\rowcolor[HTML]{eff6fc} 
\cellcolor[HTML]{FFFFFF} & Raw\_Aug\_HR & 78.21\stdvun{9.38} & 77.68\stdvun{10.30} & 79.57\stdvun{11.65} & 80.39\stdvun{8.39} & 78.96\stdvun{9.93}\\
\rowcolor[HTML]{e6f1fa} 
\cellcolor[HTML]{FFFFFF} & T2I\_Aug\_HR & 80.54\stdvun{11.71} & 80.88\stdvun{13.50} & 80.80\stdvun{12.88} & 81.69\stdvun{9.69} & 80.97\stdvun{11.94}\\
\rowcolor[HTML]{DDEBF7} 
\multirow{-5}{*}{\cellcolor[HTML]{FFFFFF}Macro} & \textbf{FLARE} (Ours) & \textbf{87.69\stdvunnn{18.86}} & \textbf{84.67\stdvunnn{17.29}} & \textbf{85.70\stdvunnn{17.78}} & \textbf{86.50\stdvunnn{14.50}} & \textbf{86.14\stdvunnn{17.11}} \\ \hline 
\rowcolor[HTML]{F2F2F2} 
\cellcolor[HTML]{FFFFFF} & Raw\_LR \cite{carrasco2019deep_raw_lr} \stdvuno{bs.} & 55.80\stdvuno{bs.} & 54.16\stdvuno{bs.} & 55.62\stdvuno{bs.} & 58.19\stdvuno{bs.} & 55.94\stdvuno{bs.} \\
\rowcolor[HTML]{f6fbff} 
\cellcolor[HTML]{FFFFFF} & Raw\_Aug\_LR \cite{perez2017effectiveness_raw_aug_lr} & 66.58\stdvun{10.78} & 67.21\stdvun{13.05} & 67.42\stdvun{11.80} & 66.43\stdvun{8.24} & 66.91\stdvun{10.97} \\
\rowcolor[HTML]{eff6fc} 
\cellcolor[HTML]{FFFFFF} & Raw\_Aug\_HR & 70.53\stdvun{14.73} & 69.89\stdvun{15.73} & 72.53\stdvun{16.91} & 72.44\stdvun{14.25} & 71.35\stdvun{15.41} \\
\rowcolor[HTML]{e6f1fa} 
\cellcolor[HTML]{FFFFFF} & T2I\_Aug\_HR & 77.19\stdvun{21.39} & 76.05\stdvun{21.89} & 77.06\stdvun{21.44} & 77.02\stdvun{18.83} & 76.83\stdvun{20.89}\\
\rowcolor[HTML]{DDEBF7} 
\multirow{-5}{*}{\cellcolor[HTML]{FFFFFF}Fine-grained} & \textbf{FLARE} (Ours) & \textbf{83.58\stdvunnn{27.78}} & \textbf{81.07\stdvunnn{26.91}} & \textbf{83.60\stdvunnn{27.98}} & \textbf{82.58\stdvunnn{24.39}} & \textbf{82.71\stdvunnn{26.77}}\\ \hline

\end{tabular}
}
\label{tab:acc_f1_macro}
\end{table}}

\begin{figure}[h]
    \centering
    \includegraphics[width=0.70\textwidth]{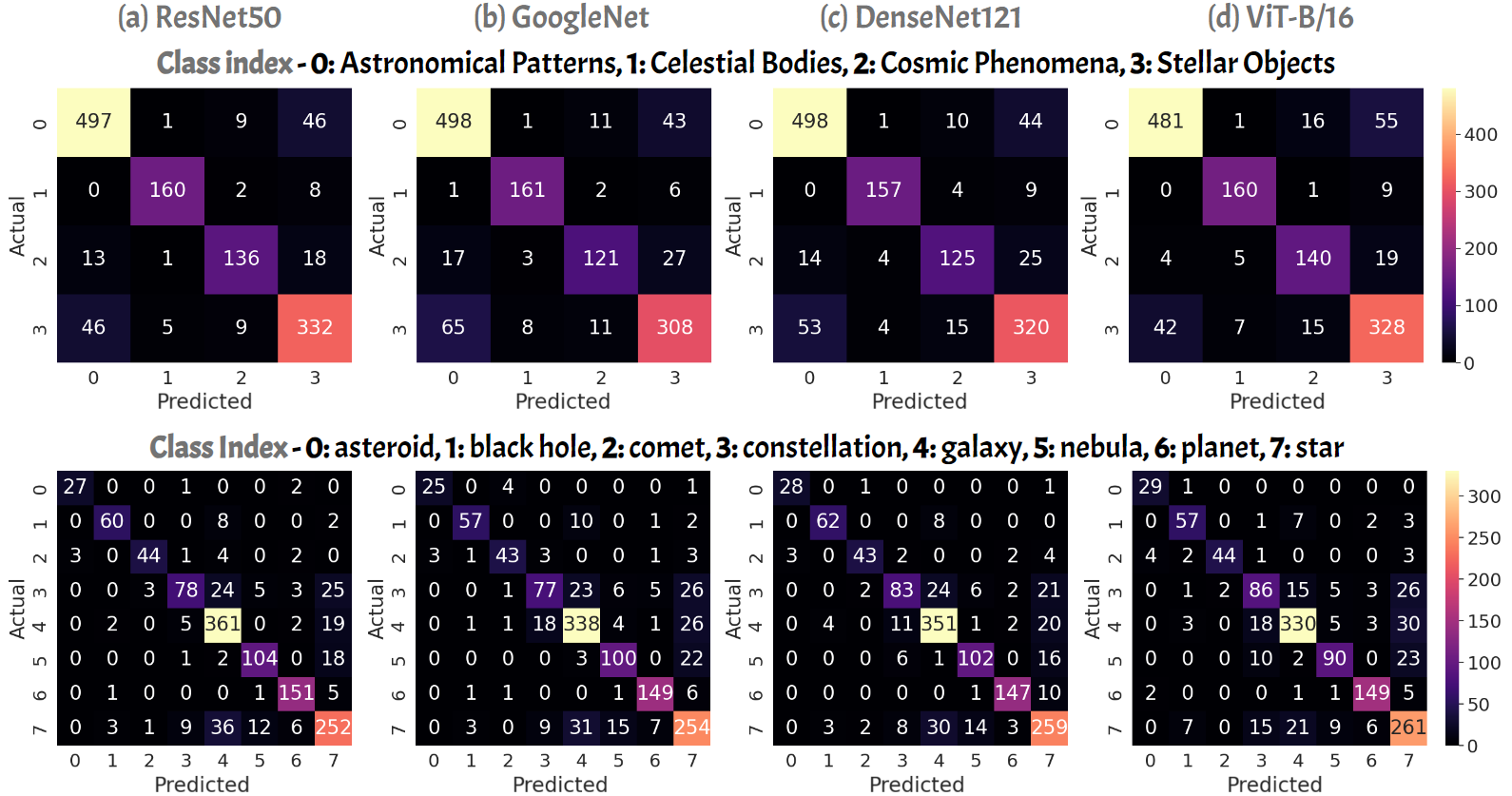}
    \caption{Confusion Matrix of 4 best models on our combined SpaceNet dataset}
    \label{fig:confusion_matrix}
\end{figure}

\begin{figure}[h]
    \centering
    \includegraphics[width=\textwidth]{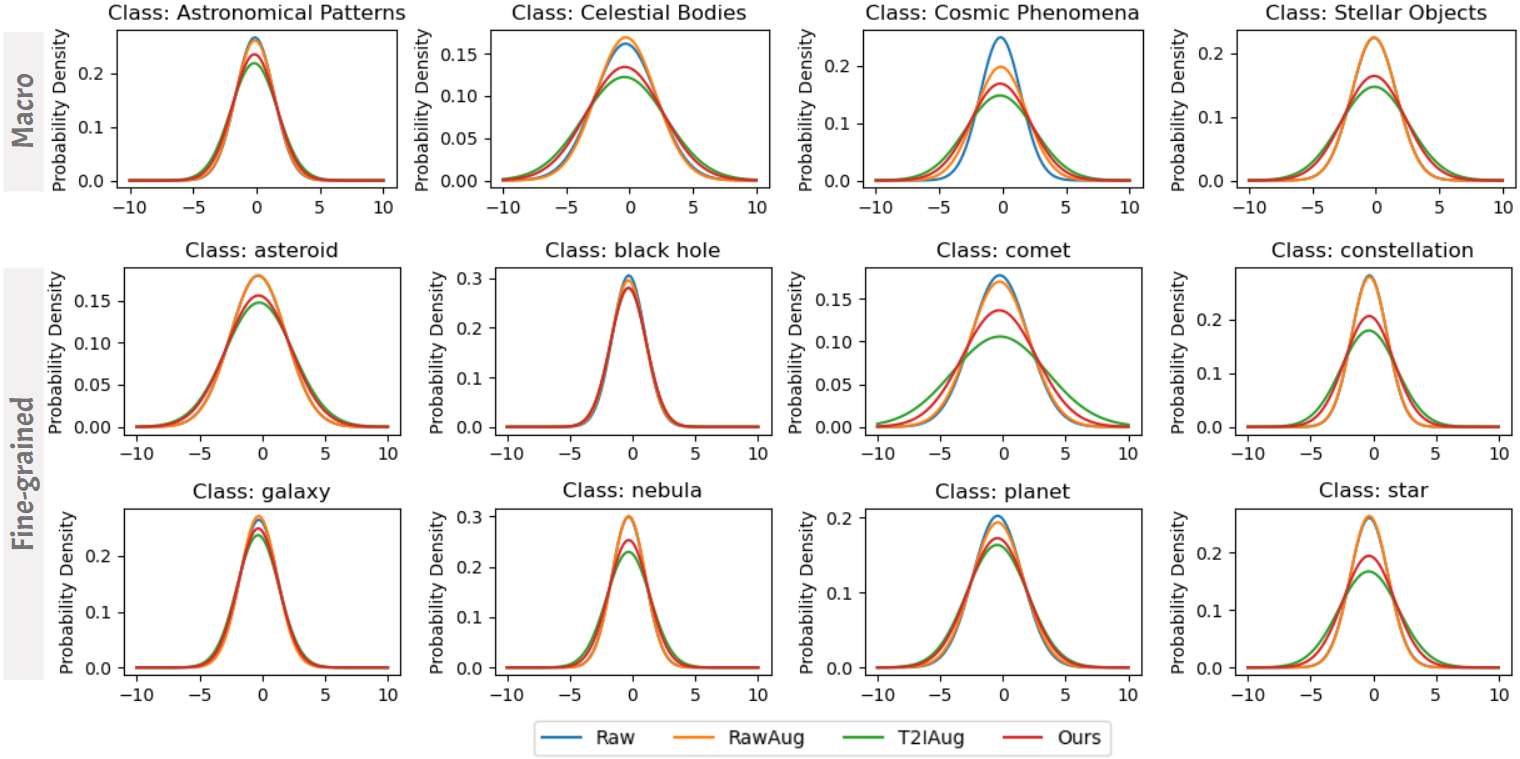}
    \caption{Normal distribution representing Mean and Variances across different dataset variants.}
    \label{fig:bell_curves}
\end{figure}

{\renewcommand{\arraystretch}{1.4}
\begin{table}[t]
\centering
\caption{Notations describing different datasets.}
\resizebox{0.75\textwidth}{!}{%
\begin{tabular}{cl}
\hline
\textbf{Notation} & \textbf{Dataset Description} \\
\hline
$D^{LR}$ & Raw\_LR, \textit{i.e.}, Raw Lower Resolution images \\
$D^{LR}_{Aug}$ & Raw\_Aug\_LR, \textit{i.e.}, Raw Lower Resolution images and augmentations \\
$D^{HR}_{Aug}$ & Raw\_Aug\_HR, \textit{i.e.}, Raw Higher Resolution images \\
$D^{HR}_{T2I}$ & T2I\_Aug\_HR, \textit{i.e.}, Syntheic samples in Higher Resolution \\
$\tilde{D}^{HR}$ & SpaceNet (Ours) \\
\hline
\end{tabular}
}
\label{tab:notations}
\end{table}
}

\end{document}